\documentclass[11pt]{article}
\usepackage{dialogue2021}

\usepackage{ifxetex}
\ifxetex
    \usepackage{fontspec}
    \setromanfont{Times New Roman}
\else
  \usepackage[T1]{fontenc}
  \usepackage[utf8]{inputenc}
  \usepackage{cmap}
  \usepackage{times}
  \usepackage{latexsym}

\fi

\usepackage[russian,british]{babel}
\usepackage{url}
\usepackage{pgf}

\usepackage{covington} %% comment it out if you do not require it

\dialogfinalcopy % Uncomment this line for the final submission

\title{Russian News Clustering and Headline Selection Shared Task}

\author{Ilya Gusev \\
  Moscow Institute of Physics and Technology \\
  Moscow, Russia \\
  {\tt ilya.gusev@phystech.edu} \\\And
  Ivan Smurov \\
  ABBYY, \\
  Moscow, Russia \\
  {\tt ivan.smurov@abbyy.com} \\}

\date{}

\begin{document}
\maketitle
\begin{abstract}
  This paper presents the results of the Russian News Clustering and Headline Selection shared task. As a part of it, we propose the tasks of Russian news event detection, headline selection, and headline generation. These tasks are accompanied by datasets and baselines. The presented datasets for event detection and headline selection are the first public Russian datasets for their tasks. The headline generation dataset is based on clustering and provides multiple reference headlines for every cluster, unlike the previous datasets. Finally, the approaches proposed by the shared task participants are reported and analyzed.
  
  
%   \textbf{DOI:} 10.28995/2075-7182-2021-20-XX-XX
\end{abstract}

\selectlanguage{russian}
\begin{center}
  \russiantitle{Дорожка по кластеризации и выбору заголовков для русских новостей}

  \medskip \setlength\tabcolsep{2cm}
  \begin{tabular}{cc}
    \textbf{Гусев И. О.} & \textbf{Смуров И. М.}\\
      МФТИ & ABBYY\\
      Москва, Россия & Москва, Россия \\
      {\tt ilya.gusev@phystech.edu} &  {\tt ivan.smurov@abbyy.com}
  \end{tabular}
  \medskip
\end{center}

\begin{abstract}
  В статье представлены результаты соревнования по кластеризации и выбору заголовков для новостей на русском. В соревновании предлагаются задачи по обнаружению новостных событий, отбору и написанию заголовков для новостей. Вместе с задачами предоставляются наборы данных и базовые решения. Представленные наборы данных для обнаружения новостных событий и выбора заголовков — первые общедоступные наборы данных на русском языке для своих задач. Набор данных для написания заголовков использует кластеризацию и содержит набор эталонных заголовков для каждого кластера, в отличие от предшественников. Представлены и проанализированы подходы, предложенные участниками соревнования.
  
  \textbf{Ключевые слова:} кластеризация, определение событий, выбор заголовков, написание заголовков, новости, эмбеддинги, русский язык, набор данных
\end{abstract}
\selectlanguage{british}

\section{Introduction}

Automatic news feeds and aggregators are a common way to read, search and analyze news. Event clustering is one of the core features for many news aggregators, including Google News and Yandex News. The news event clustering task is to collect news from different news agencies about the same event. It is essential to present all events from different perspectives to create a comprehensive picture. Moreover, selecting the most suitable headline and other entities is possible only within such clusters. The task can also be helpful for news monitoring systems. Correct clustering should help to accurately calculate any statistics about events, companies, or people's mentions in the news.

As for natural language research, this task is a good benchmark for different text clusterization or classification models. For example, the models should identify various types of paraphrasing and recognize named entities to distinguish clusters correctly.

After event clustering, it is necessary to choose the most relevant headline for every cluster. Thus, it is the headline selection task. One can define relevance in many ways. We provide a list of criteria by which a title can be considered suitable. The main points of this list are informativeness and the absence of clickbait.

Instead of choosing one of the existing headlines, one can generate a completely novel headline based on news texts. The main reason to do this is to evade the situations where all presented headlines are inadequate. 

To solve these three problems, we composed a shared task as a part of the Dialogue 2021 conference\footnote{http://www.dialog-21.ru/evaluation/}. These three problems are considered as independent tasks and evaluated separately. We provided datasets for each of them and hosted Kaggle-like competitions on the CodaLab platform to determine the best models.

The primary source of data for all tasks is the Telegram Data Clustering contest\footnote{https://contest.com/docs/data\_clustering2}. Telegram conducted it in 2020. The task was to build a news aggregator over a provided document collection. The subtasks were language detection, category detection, and event clustering. Only HTML documents without any additional annotations were given.

The contributions of our paper are as follows: we present datasets for the Russian news event clustering, headline selection, and headline generation tasks along with baselines. The first two datasets are the first of their kind for the Russian language. The last one is the first dataset for headline generation for Russian with multiple reference headlines. Furthermore, we present and analyze some of the works of the shared task participants.

\section{Related work}
Many works on event clustering exist. This task is also known as "documental event detection" or just "event detection"\cite{allan1998topic, chen2020history}. We will use these names as synonyms in this paper.

The Topic Detection and Tracking (TDT) initiative \cite{allan1998topic} was the first significant effort for this task. The event detection task was a part of TDT. There were only 25 events in the dataset. The CMU approach used TF-IDF embeddings, hierarchical agglomerative clustering with average linking, and a time window with incremental IDF for online detection.

Azzopardi et al.\cite{azzopardi2012incremental} used TF-IDF document representations and incremental k-means clustering with cosine similarity between these representations. There were no human annotations involved, and the comparison was with Google News automatic clustering.

Miranda et al.\cite{miranda-etal-2018-multilingual} focused on cross-lingual clustering. They used TF-IDF document representation with separate vectors for different document sections and incremental centroid clustering. There were also two versions of these representations: monolingual and cross-lingual. They introduced a multilingual dataset adapted from Rupnik et al.\cite{rupnik} containing articles in English, Spanish and German. It was manually annotated with monolingual and cross-lingual event labels. Linger et al.\cite{linger2020batch} utilized the same multilingual dataset, but used the multilingual DistilBERT\cite{distilbert} and the Sentence-BERT\cite{sentence_bert} triplet network structure for multilingual document representation.

There are some works on news clustering that focus on thematic news clustering instead of event clustering (with categories like "sports" or "technologies") \cite{stankevicius}.

There are few papers on Russian news event clustering. Dobrov et al.\cite{dobrov2010basic} described event clustering on a ROMIP-2006 news collection. Three different document representations were utilized: TF-IDF for texts, TF-IDF for titles, and a conceptual index. Several clustering methods were used, including hierarchical clustering, DBSCAN\cite{dbscan}, and centroid clusterings. They utilized custom software for creating a manual reference clustering.

Voropaev et al.\cite{voropaev} used the same document collection we do, Telegram Data Clustering contest document collection, but without a large-scale manual markup. They took the solution of one of the Telegram contest participants as a reference clustering. They utilized TF-IDF, BERT\cite{bert, rubert}, LASER\cite{laser} for document representation, and hierarchical clustering or DBSCAN\cite{dbscan}.

As for headline selection, we did not find any papers with a similar task definition. However, there are several works on clickbait detection in headlines\cite{potthast2016clickbait, chakraborty2016stop} which is a part of our task.

The headline generation is a well-covered task in previous papers. Statistical models for news headline generation in Banko et al.\cite{banko2000headline} are among the first approaches to this task. Takase et al.\cite{takase2016neural} were the first to use encoder-decoder neural models for headline generation. In 2019, there was a Dialogue Evaluation shared task for Russian news headline generation\cite{headline_gen}. The main drawback of the 2019 shared task was the fact that it was about single-document headline generation. It means that there was only one actual headline for every article, so automatic evaluation methods were not effective in this setup.

\section{Clustering}
\subsection{Data and metrics}

For the first two tasks, we took news documents from the Telegram Data Clustering contest covering three days of May 2020. After using all baseline clustering algorithms, we sampled a set of document pairs from their output and included a small number of random pairs. Next, we annotated every pair with Yandex Toloka, a Russian crowdsourcing platform. The task was to determine whether two documents describe the same event. The annotators were provided with a comprehensive guide and a simple heuristic: "if two titles of two documents are interchangeable, these documents are from the same cluster". Five people annotated each pair. Annotators were required to pass training, exam, and their work was continuously evaluated through the control pairs ("honeypots"). We included only pairs with an agreement of 4/5 or 5/5 in the final markup.

\begin{table}[htbp]\label{tab1}
\centering
\begin{tabular}{|c|c|c|c|}\hline
& May 25 & May 27 & May 29 \\\hline\hline
\#Documents & 19380 &  20080 & 19096 \\\hline
\#Pairs & 14838 & 8493 & 8476 \\\hline
\#Positives & 7301 & 3918 & 3896 \\ \hline
Fraction of positives & 49.2\% & 46.1\% & 46.0\% \\ \hline
\end{tabular}
\caption{Clustering dataset statistics}
\label{tabDocStat}
\end{table}

\begin{figure}[htbp]
% \centering
\centerline{\includegraphics[scale=.35]{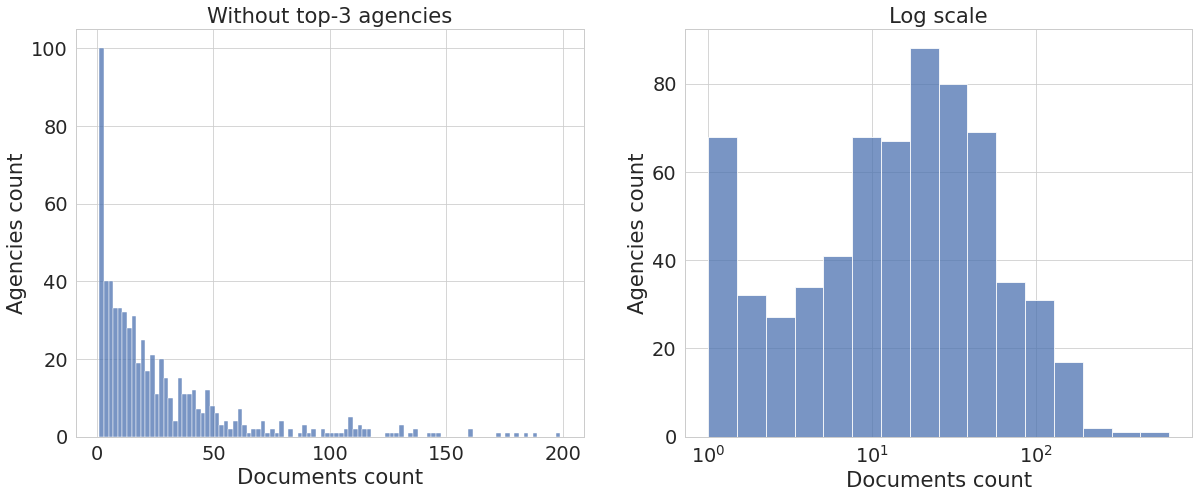}}
\caption{News agencies distribution by the number of documents, May 25}
\label{figDistrib}
\vspace{-1.2cm}
\end{figure}

The final statistics for every day are shown in Table~\ref{tabDocStat}. There are 661 unique hosts in the training document collection and at least 500 unique news agencies. The top 5 news agencies by the number of documents are shown in Table~\ref{tabTopHosts}. The distribution of news agencies by the number of documents is in Figure~\ref{figDistrib}.

\begin{table}[htbp]\label{tab2}
\centering
\begin{tabular}{|c|c|}\hline
Host & \#Documents \\\hline\hline
tass.ru & 645 \\\hline
lenta.ru & 301 \\\hline
www.mk.ru & 223 \\ \hline
www.rosbalt.ru & 199 \\ \hline
ura.news & 188 \\ \hline
\end{tabular}
\caption{Top 5 hosts by the number of documents, May 25}
\label{tabTopHosts}
\end{table}

According to annotation guidelines, a pair of documents refer to the same cluster when they have the same: time of the event; numbers, such as the stock price of a company or the number of victims; locations. A pair of documents are not from the same cluster when they contain: inconsistent facts, such as the time or place of the event or significantly distinguished number of victims; description of an event in one of the documents, and a commentary on this event in another document. These criteria are similar to the definition of an event in TDT\cite{allan1998topic}.

We do not publish documents themselves, only two URLs and an annotation result, to evade legal issues. However, Telegram shares archives with all documents, so it is easy to join documents and annotations. An example of the joining process can be found in the baseline script\footnote{https://github.com/dialogue-evaluation/Russian-News-Clustering-and-Headline-Generation/blob/main/baselines.ipynb}. No additional scraping is needed.

In the shared task rules, we forbade the use of any news documents from the test set for training or pretraining, including pretraining word vectors on these documents. In the real world, texts are not available for the model in advance, so the tested models must not use the test-time texts in any way. However, it was permissible to use texts with earlier dates from the Telegram contest for pretraining.

It is possible to collect non-pairwise annotations for clustering with crowdsourcing methods, but it is hard to organize proper quality control of the annotation process. Non-pairwise annotations we know of for event detection were done by experts, not via crowdsourcing.

As the primary metric for this task, we chose F1-score (corresponds to Dice similarity coefficient for clustering). The proper clustering metrics like Adjusted Rand Index\cite{rand} are not directly applicable to our dataset as we have no reference clustering for all documents. Several dozen possible pair-counting based clustering metrics\cite{clustering_indices} are available, and it is an active research question what metric is better\cite{cluster_measures}. As for our choice, we decided that the metric should be interpretable from the classification perspective, and our classes are imbalanced, so we chose F1.

\subsection{Baselines}
As a baseline clustering algorithm, we utilized hierarchical agglomerative clustering with average linking.

\begin{figure}[htbp]
\centering
\includegraphics[scale=.13]{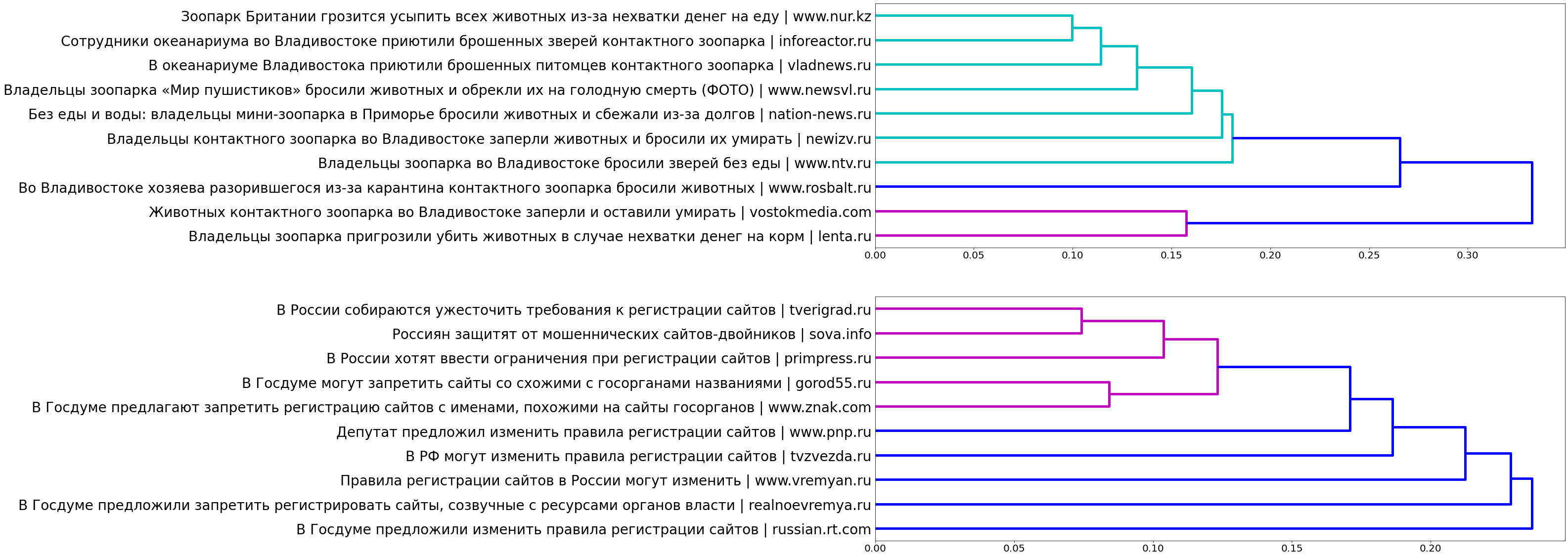}
\caption{Dendrograms for two clusters based on USE embeddings, cosine similarity, and average linking. These clusters contain several errors.}
\label{figDendogram}
\vspace{-0.1cm}
\end{figure}

We used different pretrained embeddings with cosine similarity to compute the distance matrix. The simplest ones are FastText\cite{fasttext} text embeddings, TF-IDF with truncated SVD for dimensionality reduction (latent semantic analysis, LSA\cite{lsa}), and Universal Sentence Encoder (USE\cite{use}). FastText text embeddings are computed as a concatenation of average, maximum, and minimum of FastText word embeddings. The examples of clusters based on USE embeddings are shown in Figure~\ref{figDendogram}.

\begin{figure}[htbp]
\centering
\includegraphics[scale=.12]{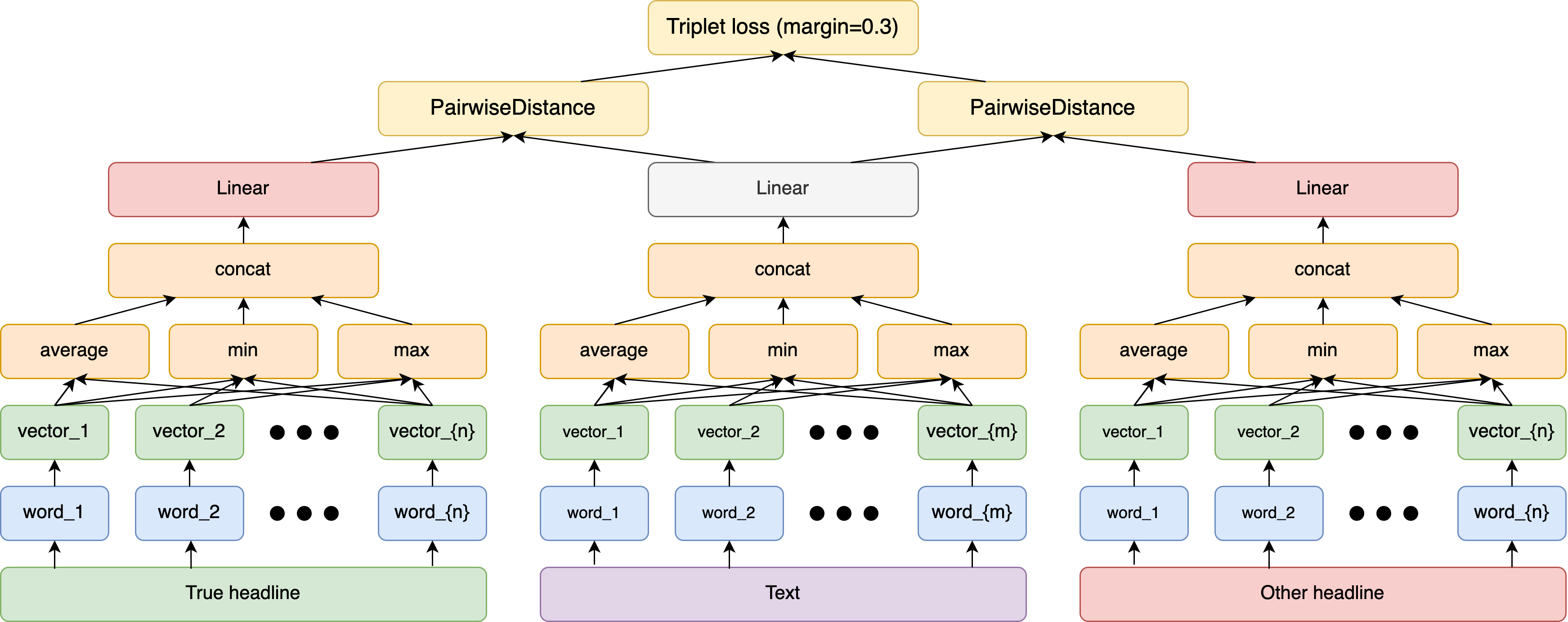}
\caption{Text2Title model architecture}
\label{figTT}
\vspace{-0.1cm}
\end{figure}

Text2Title embeddings are FastText embeddings with an additional linear matrix trained to determine whether a headline and a text are from the same document. HNSW\cite{hnsw} index on the original FastText embeddings was used to mine hard negatives for the triplet loss\cite{triplet_loss}. The model architecture is depicted on Figure~\ref{figTT}. The main advantage of this model is its speed, as it consists of only one linear layer on top of the FastText embedding.

TTGenBottleneck is a BERT seq2seq model based on RuBERT\cite{rubert} with practically disabled cross-attention between encoder and decoder. It was trained to predict a headline for a document. "Bottleneck" is the encoder embedding of the first token: the only embedding decoder can attend. We design it\footnote{https://huggingface.co/IlyaGusev/gen\_title\_tg\_bottleneck\_encoder} to contain all information from the text needed to generate a headline.

We emphasize that none of the baseline models were fine-tuned to the news event detection task in any way.

We present the results of all baselines in Table~\ref{tabClusteringBaselines}. Text2Title has better scores than plain FastText, as additional pretraining helps to create more representative embeddings. TTGenBottleneck is the best baseline model.

\begin{table}[htbp]\label{tab3}
\centering
\begin{tabular}{|c|c|c|c|}\hline

Model & Validation & Public LB & Private LB \\\hline\hline
TTGenBottleneck & 93.4 & 93.9 & 93.7 \\ \hline
% RuBERT-MLM* &  &  &  \\ \hline
USE & 89.3 & 89.4 & 87.8 \\ \hline
Text2Title & 86.5 & 86.4 & 84.8 \\\hline
LSA & 83.1 & 82.7 & 80.5 \\ \hline
FastText & 80.9 & 81.9 & 80.1 \\\hline
\end{tabular}
\caption{F1-scores for positive pairs in \% for baseline models}
\label{tabClusteringBaselines}
\end{table}

\subsection{Results}
The task can be solved both as a classification task or as a clustering task. The baselines were for the clustering only, but many participants preferred to use classification models.

The final results are in Table~\ref{tabClusteringResults}. There are several participants without scores in the private leaderboard. We made a technical mistake that disabled automatic score evaluation of the private part, and the participants had to resubmit their answers. Not all of them managed to do that.

\begin{table}[htbp]\label{tab4}
\centering
\begin{tabular}{|c|c|c|c|c|}\hline
Rank & Codalab login & Public LB & Private LB \\ \hline\hline
\textbf{1} & \textbf{maelstorm} \footnotemark & \textbf{96.9} & \textbf{96.0} \\ \hline
2 & naergvae \cite{participants135} & 96.7 & 96.0 \\ \hline
3 & g2tmn \cite{participants131} \footnotemark & 96.5 & 95.7 \\ \hline
4 & Kouki \cite{participants136} & 95.5 & 95.5 \\ \hline
5 & alexey.artsukevich & 95.8 & 95.3 \\ \hline
6 & smekur \cite{participants136}  & 94.6 & 93.9 \\ \hline
7 & nikyudin & 93.8 & 93.0 \\ \hline
8 & landges & 91.6 & 90.6 \\ \hline
9 & kapant & 90.7 & 89.9 \\ \hline
10 & bond005 & 90.2 & 89.2 \\ \hline
11 & anonym & 90.6 & 89.1 \\ \hline
12 & mashkka\_t \cite{participants138} & 85.3 & 71.5 \\ \hline
13 & vatolinalex \cite{participants136}  & 95.2 & 47.6  \\ \hline
- & blanchefort & 94.1 &  \\ \hline
- & imroggen & 90.3 & \\ \hline
- & Abiks & 89.4 & \\ \hline
- & dinabpr \cite{participants138} & 84.4 & \\ \hline
\end{tabular}
\caption{Final results of the clustering track, F1-scores for positive pairs in \%}
\label{tabClusteringResults}
\end{table}

\footnotetext[5]{Leonid Pugachev and Alim Adelshin, DeepPavlov}
\footnotetext[6]{https://github.com/oldaandozerskaya/DE2021\_news\_similarity}

The best solution was a classification model, an ensemble of 4 bert-base-multilingual models trained with stochastic weight averaging (SWA)\cite{izmailov2018averaging}. The second-placed model was also classification-based and consisted of a single RuBERT\cite{rubert} in a standard pair classification setting. Finally, the third-placed model was a hard-voting classification ensemble of RuBERT models.

The best clustering-based model (4th place overall) was based on SBERT\cite{sentence_bert}: a siamese BERT model over RuBERT. The authors\cite{participants136} improved the model by using Global Multihead Pooling\cite{chen-etal-2018-enhancing} and contrastive loss. It should be noted that this model scored only 0.005 F1 less than the best classification model on the private test. Additionally, this is the only model that did not experience score deduction on the private test set compared to the public set. 

Unsurprisingly all top-placed models utilized pretrained masked language models. In particular, the top three classification-based models and the best clustering-based model (4th place overall) all used BERT\cite{bert}. The best solution not utilizing language model embeddings was able to score 0.930 F1 (7th place overall) and had the following architecture: CatBoost\cite{catboost} over FastText\cite{fasttext} and USE\cite{use} as well as some handcrafted features (hyperlinks and named entities intersections).

Another system of note is GPT-3-based\cite{gpt3} zero-shot model. The idea is as follows: the two headlines' perplexity is computed and compared to thresholds. While the final score is rather unimpressive, 0.7 F1, the completely unsupervised nature of training makes this result worth mentioning. 

\pagebreak
\section{Headline selection}
\subsection{Data and metrics}
The documents for this task are from the same collection as the documents for the clustering task. The task was, for each given pair of headlines, to predict which headline is better. There are four possible options: "left", "right", "draw", and "bad". The "bad" option is for the case if two headlines are from different news events. Annotation conditions are the same as for the clustering task. According to annotation guidelines, one headline is better than the other if some of the following conditions are met: it contains more information than the other; it does not hide any details; it does not contain undefined entities; it has no grammatical errors; it is not emotional; it is not too wordy. Some of these criteria are subjective, so we use only examples with high agreement to reduce subjectivity.

The final statistics for every day are shown in Table~\ref{tabSelectionStats}.

\begin{table}[htbp]\label{tab5}
\centering
\begin{tabular}{|c|c|c|c|}\hline
& May 25 & May 27 & May 29 \\\hline\hline
\#Pairs & 5091 & 3147 & 3103 \\\hline
\#Left won & 2185 & 1254 & 1216 \\ \hline
\#Right won & 2167 & 1269 & 1207 \\ \hline
\#Draw & 362 & 184 & 161 \\ \hline
\#Bad & 377 & 440 & 518 \\ \hline
\end{tabular}
\caption{Headline selection dataset statistics}
\label{tabSelectionStats}
\end{table}

The primary metric for this task is weighted accuracy. The dataset contains a lot of "bad" examples. They were added to the dataset to prevent data leaks for the clustering task. We ignore them in the denominator and numerator. The systems can predict any label for them. For the other three labels, scores in the numerator are calculated as stated in Table~\ref{tabSelectionMetrics}. We chose such weights because incorrect predictions for the "draw" pairs should not be penalized as hard as "left/right" errors. We do not provide F1-scores, as the dataset is balanced, and the only information these scores add to accuracy is how well "draw" pairs are detected.

\begin{table}[htbp]\label{tab6}
\centering
\begin{tabular}{|c|c|c|c|}\hline
 & Left  & Right & Draw\\\hline
Left & 1.0  & 0.0 & 0.5\\\hline
Right & 0.0 & 1.0 & 0.5\\\hline
Draw & 0.5 & 0.5 & 1.0\\\hline
\end{tabular}
\caption{Predictions weights in the accuracy metric}
\label{tabSelectionMetrics}
\end{table}

\subsection{Baseline}
We used a ranking gradient boosting model (CatBoost\cite{catboost}) with USE embeddings as features and PairLogit loss as a baseline for this task. This CatBoost mode is designed to take pairs as input, so it fits the task perfectly. We present the result accuracy of this model in Table~\ref{tabSelectionBaselines}.
$$PairLogitLoss(pairs, res) = -\sum_{p,n \in pairs}{ log(\frac{1}{1 + e^{-(res_p - res_n)}})}$$  

\begin{table}[htbp]\label{tab7}
\centering
\begin{tabular}{|c|c|c|c|}\hline
Model & Validation & Public LB & Private LB \\\hline\hline
USE + Catboost & 81.0 $\pm$ 1.5 & 81.2 $\pm$ 0.5  & 81.1 $\pm$ 0.4 \\\hline
\end{tabular}
\caption{Weighted accuracy for the baseline model, 5 runs, in \%}
\label{tabSelectionBaselines}
\end{table}

\subsection{Results}
Most of the successful submissions are based on the same schema as the baseline. The results are presented in Table~\ref{tabSelectionResults}. There is only one submission that surpasses the baseline significantly. It utilizes an ensemble of ranking models trained on different embeddings, including USE, SBERT\cite{sentence_bert}, RuBERT\cite{rubert}, XLM-R\cite{xlm_r}, and mT5\cite{mt5}. The most effective single model was mT5.

\begin{table}[htbp]\label{tab8}
\centering
\begin{tabular}{|c|c|c|c|}\hline
Rank & Codalab login & Public LB & Private LB\\ \hline\hline
\textbf{1} & \textbf{sopilnyak} \cite{participants140} & \textbf{86.0} & \textbf{85.4} \\ \hline
2 & landges & 81.3 & 82.0 \\ \hline
3 & nikyudin & 83.2 & 81.6 \\ \hline
4 & LOLKEK & 80.8 & 81.4 \\ \hline
5 & maelstorm & 81.8 & 79.8 \\ \hline
6 & a.korolev & 65.8 & 66.2 \\ \hline
\end{tabular}
\caption{Final results of the headline selection track, weighted accuracy in \%}
\label{tabSelectionResults}
\end{table}

\section{Headline generation}
\subsection{Data and metrics}
In this track, the task was to generate a headline for a news cluster. It should be similar to any of the headlines of cluster documents. Other formulations to this task are possible. For example, there is a much more complicated task formulation where one should generate a headline similar to the best headline of the cluster. However, we chose the weak formulation as it does not require any annotation. We were not able to use the documents and annotations of the first two tracks because of possible data leaks.

To make a dataset entirely unknown for participants, we scraped news documents from a test version of the Telegram news aggregator\footnote{https://1398.topnews.com/ru/} from March 9 to March 12, 2021. We used the TTGenBottleneck clustering method with a distance threshold skewed into precision to the detriment of recall, as it forms very restricted clusters.

We made available for participants only clustered document texts without any headlines or additional meta information. There are 6726 documents and 1035 clusters in the dataset.

This formulation can be considered more robust than the single document (and single news agency \cite{headline_gen}) headline generation as it does not force the model to generate a headline in a particular style to get good automatic metrics.

We use traditional automatic metrics for headline generation, ROUGE\cite{rouge}, and BLEU\cite{bleu}. We calculate metrics between a predicted headline and every actual headline of a cluster and use the maximum score as a prediction score.

\subsection{Baselines}
There are two baselines for this track. The first is Random Lead-1, where we choose the first sentence of a random document from the cluster as a baseline. The second baseline is a generative sequence to sequence\cite{seq2seq} RuBERT\cite{rubert} trained on news text-title pairs from the Telegram contest\footnote{https://huggingface.co/IlyaGusev/rubert\_telegram\_headlines}. We generate a headline for every text in the cluster and choose randomly from them. We present the metrics for the baselines in Table~\ref{tabGenBaseline}.

\begin{table}[htbp]\label{tab9}
\centering
\begin{tabular}{|c|c|c|c|c|c|}\hline
Model & ROUGE & BLEU & ROUGE-1 & ROUGE-2 & ROUGE-L \\\hline\hline
Seq2seq RuBERT & 44.9 $\pm$ 0.3 & 74.5 $\pm$ 0.5 & 52.5 $\pm$ 0.4 & 32.8 $\pm$ 0.4 & 49.5 $\pm$ 0.5 \\\hline
Random Lead-1 & 30.2 $\pm$ 0.5& 47.8 $\pm$ 0.4 & 36.8 $\pm$ 0.6 & 20.4 $\pm$ 0.4 & 33.6 $\pm$ 0.6 \\\hline
\end{tabular}
\caption{Results of headline generation baseline models, 5 runs, in \%}
\label{tabGenBaseline}
\end{table}

\subsection{Results}
There were only two participants in this track, mainly because of hard time restrictions. Their scores can be seen in Table~\ref{tabGenResults}. The baseline remained unbeaten.

The solution by \texttt{Rybolos} was based on fine-tuning the ruGPT-3 Large model. We suppose their metrics are low compared to baselines because of possible technical error or only a tiny part of the dataset used for training.

\begin{table}[htbp]\label{tab10}
\centering
\begin{tabular}{|c|c|c|c|c|c|}\hline
Codalab login & ROUGE & BLEU & ROUGE-1 & ROUGE-2 & ROUGE-L \\\hline\hline
LOLKEK & 38.7 & 69.5 & 46.3 & 26.4  & 43.3 \\\hline
Rybolos & 29.2 & 59.6 & 36.5 & 17.6 & 33.5 \\\hline
\end{tabular}
\caption{Final results of the headline generation track, in \%}
\label{tabGenResults}
\end{table}

\section{Conclusion}
\subsection{Reproducibility}
All materials of the shared task are available at the official repository\footnote{https://github.com/dialogue-evaluation/Russian-News-Clustering-and-Headline-Generation}, including all data, annotation guidelines, baselines, and links to the CodaLab competitions. The competitions themselves will be permanently open in order to make comparison easier for new researchers.

\subsection{Organization notes}
The timeline was as follows:
\begin{itemize}
    \item February 8: Clustering task started on Codalab.
    \item February 26: Headline selection task started on Codalab.
    \item March 13: Headline generation task started on Codalab.
    \item March 22: Final deadline for all competitions.
    \item March 28: Final deadline for paper submission.
\end{itemize}

The main reason for a short time window for submissions for the last task was that it was unclear where to get previously unseen data from roughly the same domain. We should have solved this question even before the shared task was announced. Unfortunately, the news clustering task was the only one that we fully prepared before the start.

The other major problem was with the private leaderboard for the clustering task, as the initial private part of the test dataset was a copy of a public part by our technical mistake, so it was impossible to measure metrics without resubmitting answers.

\subsection{General conclusions}
In the event detection task, most of the successful models were classification-based BERT models. However, it turns out clustering embeddings can be almost as effective when trained with correct pooling and loss function. Moreover, they generalize better, they are easier to deploy, and they are more computationally effective. It is arguably the most important takeaway of the shared task.

Unsurprisingly, big multilingual models and ensembles showed the best metrics in the headline selection task. Nevertheless, the task participants did not diverge much from the baseline solution, so we hope more sophisticated schemas and models will be developed in the future.

There was only one week to develop a valid generative model in the headline generation task. Only two participants managed to present a complete model in these challenging time restrictions, so we consider this track results inconclusive. However, the dataset itself can be helpful in future research.

\subsection{Future research}
There are several possible directions for future research:
\begin{enumerate}
\item Finding models with good accuracy/speed trade-offs is a very promising research path. Almost all of the models used by participants were extremely parameter-heavy and slow. Distillation of these models into lightweight ones is needed in order to enable their use in production systems.
\item Different clustering methods should be inspected. Solutions for the clustering task were agglomerative-centered — almost no one used BIRCH clustering\cite{zhang1996birch}, and only a few people used DBSCAN\cite{dbscan} and its modifications.
\item The Telegram Data Clustering news documents are multilingual, so datasets and models should be multilingual too. We created only a Russian dataset because we used a Russian crowdsourcing platform and had a constrained money budget. There are no principal reasons why these datasets should not be multilingual.
\item The existing clustering dataset is focused on 24-hour time windows. Clustering on more extensive periods can be significantly harder. Moreover, online event detection (as it originally was in TDT) is also possible and requires windowed or incremental clustering.
\item Synthetic checklists\cite{checklists} can be used to evaluate the quality of event detection and headline selection. It is easy to create pairs of documents differing only in numbers, entities, or time of an event.
\item As for the headline generation task, the clustering enables proper conditioning. It is possible to train a model to write headlines in the style of the particular news agency without a thematic bias, as agencies will be equally presented in clusters.
\item It is possible to revisit other TDT tasks in light of recent advances in building document representations with big pretrained models like BERT.
\end{enumerate}

\subsubsection*{Acknowledgements}
We would like to thank the participants of all three tracks, especially Tatiana Shavrina, Ivan Bondarenko, and Nikita Yudin for helpful comments and valuable suggestions. 

\color{blue}\section*{References}

\makeatletter
\renewcommand{\section}{\@gobbletwo}
\makeatother
\bibliography{dialogue.bib}

\begin{thebibliography}{10}
\def\selectlanguageifdefined#1{
\expandafter\ifx\csname date#1\endcsname\relax
\else\selectlanguage{#1}\fi}
\providecommand*{\href}[2]{{\small #2}}
\providecommand*{\url}[1]{{\small #1}}
\providecommand*{\BibUrl}[1]{\url{#1}}
\providecommand{\BibAnnote}[1]{}
\providecommand*{\BibEmph}[1]{#1}
\ProvideTextCommandDefault{\cyrdash}{\iflanguage{russian}{\hbox
  to.8em{--\hss--}}{\textemdash}}
\providecommand*{\BibDash}{\ifdim\lastskip>0pt\unskip\nobreak\hskip.2em plus
  0.1em\fi
\cyrdash\hskip.2em plus 0.1em\ignorespaces}
\renewcommand{\newblock}{\ignorespaces}

\bibitem{laser}
\selectlanguageifdefined{english}
\BibEmph{Artetxe~Mikel, Schwenk~Holger}. Massively Multilingual Sentence
  Embeddings for Zero-Shot Cross-Lingual Transfer and Beyond~// \BibEmph{Trans.
  Assoc. Comput. Linguistics}. \BibDash
\newblock 2019. \BibDash
\newblock Vol.~7. \BibDash
\newblock P.~597--610. \BibDash
\newblock Access mode:
  \BibUrl{https://transacl.org/ojs/index.php/tacl/article/view/1742}.

\bibitem{izmailov2018averaging}
\selectlanguageifdefined{english}
Averaging weights leads to wider optima and better generalization~/
  Pavel~Izmailov, Dmitrii~Podoprikhin, Timur~Garipov et~al.~// \BibEmph{arXiv
  preprint arXiv:1803.05407}. \BibDash
\newblock 2018.

\bibitem{azzopardi2012incremental}
\selectlanguageifdefined{english}
\BibEmph{Azzopardi~Joel, Staff~Christopher}. Incremental Clustering of News
  Reports~// \href{http://dx.doi.org/10.3390/a5030364}{\BibEmph{Algorithms}}.
  \BibDash
\newblock 2012. \BibDash
\newblock Vol.~5, no.~3. \BibDash
\newblock P.~364--378. \BibDash
\newblock Access mode: \BibUrl{https://doi.org/10.3390/a5030364}.

\bibitem{participants135}
\selectlanguageifdefined{english}
{BERT for Russian news clustering}~/ ~Khaustov, ~Gorlova, ~Kalmykov, ~Kabaev~//
  Computational Linguistics and Intellectual Technologies: Papers from the
  Annual Conference ``Dialogue''. \BibDash
\newblock Vol.~XX. \BibDash
\newblock 2021. \BibDash
\newblock P.~xx--xx.

\bibitem{banko2000headline}
\selectlanguageifdefined{english}
\BibEmph{Banko~Michele, Mittal~Vibhu~O, Witbrock~Michael~J}. Headline
  generation based on statistical translation~// Proceedings of the 38th Annual
  Meeting of the Association for Computational Linguistics. \BibDash
\newblock 2000. \BibDash
\newblock P.~318--325.

\bibitem{bert}
\selectlanguageifdefined{english}
Bert: Pre-training of deep bidirectional transformers for language
  understanding~/ Jacob~Devlin, Ming-Wei~Chang, Kenton~Lee,
  Kristina~Toutanova~// \BibEmph{arXiv preprint arXiv:1810.04805}. \BibDash
\newblock 2018.

\bibitem{checklists}
\selectlanguageifdefined{english}
\href{http://dx.doi.org/10.18653/v1/2020.acl-main.442}{Beyond Accuracy:
  Behavioral Testing of {NLP} Models with {C}heck{L}ist}~/ Marco~Tulio~Ribeiro,
  Tongshuang~Wu, Carlos~Guestrin, Sameer~Singh~// Proceedings of the 58th
  Annual Meeting of the Association for Computational Linguistics. \BibDash
\newblock Online~: Association for Computational Linguistics, 2020. \BibDash
  Jul. \BibDash
\newblock P.~4902--4912. \BibDash
\newblock Access mode:
  \BibUrl{https://www.aclweb.org/anthology/2020.acl-main.442}.

\bibitem{bleu}
\selectlanguageifdefined{english}
\href{http://dx.doi.org/10.3115/1073083.1073135}{{B}leu: a Method for Automatic
  Evaluation of Machine Translation}~/ Kishore~Papineni, Salim~Roukos,
  Todd~Ward, Wei-Jing~Zhu~// Proceedings of the 40th Annual Meeting of the
  Association for Computational Linguistics. \BibDash
\newblock Philadelphia, Pennsylvania, USA~: Association for Computational
  Linguistics, 2002. \BibDash Jul. \BibDash
\newblock P.~311--318. \BibDash
\newblock Access mode: \BibUrl{https://www.aclweb.org/anthology/P02-1040}.

\bibitem{catboost}
\selectlanguageifdefined{english}
CatBoost: unbiased boosting with categorical features~/ Liudmila~Prokhorenkova,
  Gleb~Gusev, Aleksandr~Vorobev et~al.~// \BibEmph{arXiv preprint
  arXiv:1706.09516}. \BibDash
\newblock 2017.

\bibitem{chen-etal-2018-enhancing}
\selectlanguageifdefined{english}
\BibEmph{Chen~Qian, Ling~Zhen-Hua, Zhu~Xiaodan}. Enhancing Sentence Embedding
  with Generalized Pooling~// Proceedings of the 27th International Conference
  on Computational Linguistics. \BibDash
\newblock Santa Fe, New Mexico, USA~: Association for Computational
  Linguistics, 2018. \BibDash Aug. \BibDash
\newblock P.~1815--1826. \BibDash
\newblock Access mode: \BibUrl{https://www.aclweb.org/anthology/C18-1154}.

\bibitem{potthast2016clickbait}
\selectlanguageifdefined{english}
Clickbait detection~/ Martin~Potthast, Sebastian~K{\"o}psel, Benno~Stein,
  Matthias~Hagen~// European Conference on Information Retrieval~/ Springer.
  \BibDash
\newblock 2016. \BibDash
\newblock P.~810--817.

\bibitem{dbscan}
\selectlanguageifdefined{english}
A Density-Based Algorithm for Discovering Clusters in Large Spatial Databases
  with Noise~/ Martin~Ester, Hans-Peter~Kriegel, Jörg~Sander, Xiaowei~Xu~//
  Proceedings of the Second International Conference on Knowledge Discovery and
  Data Mining (KDD-96). \BibDash
\newblock AAAI Press, 1996. \BibDash
\newblock P.~226--231.

\bibitem{distilbert}
\selectlanguageifdefined{english}
DistilBERT, a distilled version of BERT: smaller, faster, cheaper and lighter~/
  Victor~Sanh, Lysandre~Debut, Julien~Chaumond, Thomas~Wolf~// \BibEmph{arXiv
  preprint arXiv:1910.01108}. \BibDash
\newblock 2019.

\bibitem{dobrov2010basic}
\selectlanguageifdefined{english}
\BibEmph{Dobrov~Boris, Pavlov~Andrey}. Basic line for news clusterization
  methods evaluation~// Proceedings of the 5-th Russian Conference RCDL-2010.
  \BibDash
\newblock 2010.

\bibitem{fasttext}
\selectlanguageifdefined{english}
Enriching word vectors with subword information~/ Piotr~Bojanowski,
  Edouard~Grave, Armand~Joulin, Tomas~Mikolov~// \BibEmph{Transactions of the
  Association for Computational Linguistics}. \BibDash
\newblock 2017. \BibDash
\newblock Vol.~5. \BibDash
\newblock P.~135--146.

\bibitem{participants131}
\selectlanguageifdefined{english}
\BibEmph{Glazkova~Anna}. {Towards News Aggregation in Russian: a BERT-based
  Approach to News Article Similarity Detection}~// Computational Linguistics
  and Intellectual Technologies: Papers from the Annual Conference
  ``Dialogue''. \BibDash
\newblock Vol.~XX. \BibDash
\newblock 2021. \BibDash
\newblock P.~xx--xx.

\bibitem{cluster_measures}
\selectlanguageifdefined{english}
\BibEmph{G{\"o}sgens~Martijn, Tikhonov~Alexey, Prokhorenkova~Liudmila}.
  Systematic Analysis of Cluster Similarity Indices: How to Validate Validation
  Measures~// \BibEmph{arXiv preprint arXiv:1911.04773}. \BibDash
\newblock 2019.

\bibitem{clustering_indices}
\selectlanguageifdefined{english}
Ground Truth Bias in External Cluster Validity Indices~/ Yang~Lei,
  James~C.~Bezdek, Simone~Romano et~al.~//
  \href{http://dx.doi.org/10.1016/j.patcog.2016.12.003}{\BibEmph{Pattern
  Recogn.}} \BibDash
\newblock 2017. \BibDash May. \BibDash
\newblock Vol.~65, no.~C. \BibDash
\newblock P.~58–70. \BibDash
\newblock Access mode: \BibUrl{https://doi.org/10.1016/j.patcog.2016.12.003}.

\bibitem{chen2020history}
\selectlanguageifdefined{english}
A History and Theory of Textual Event Detection and Recognition~/ Yanping~Chen,
  Zehua~Ding, Qinghua~Zheng et~al.~// \BibEmph{IEEE Access}. \BibDash
\newblock 2020. \BibDash
\newblock Vol.~8. \BibDash
\newblock P.~201371--201392.

\bibitem{triplet_loss}
\selectlanguageifdefined{english}
\BibEmph{Hoffer~Elad, Ailon~Nir}. Deep metric learning using Triplet
  network.~// ICLR (Workshop)~/ Ed.\ by\ Yoshua~Bengio, Yann~LeCun. \BibDash
\newblock 2015. \BibDash
\newblock Access mode:
  \BibUrl{http://dblp.uni-trier.de/db/conf/iclr/iclr2015w.html#HofferA14}.

\bibitem{lsa}
\selectlanguageifdefined{english}
Indexing by latent semantic analysis.~/ S.~Deerwester, S.T.~Dumais, G.W.~Furnas
  et~al.~// \BibEmph{Journal of the American Society for Information Science
  41}. \BibDash
\newblock 1990. \BibDash
\newblock P.~391--407.

\bibitem{rubert}
\selectlanguageifdefined{english}
\BibEmph{Kuratov~Yuri, Arkhipov~Mikhail}. Adaptation of Deep Bidirectional
  Multilingual Transformers for Russian Language~// \BibEmph{CoRR}. \BibDash
\newblock 2019. \BibDash
\newblock Vol. abs/1905.07213. \BibDash
\newblock \href{http://arxiv.org/abs/1905.07213}{1905.07213}.

\bibitem{gpt3}
\selectlanguageifdefined{english}
Language Models are Few-Shot Learners~/ Tom~Brown, Benjamin~Mann, Nick~Ryder
  et~al.~// Advances in Neural Information Processing Systems. \BibDash
\newblock Vol.~33. \BibDash
\newblock Curran Associates, Inc., 2020. \BibDash
\newblock P.~1877--1901. \BibDash
\newblock Access mode:
  \BibUrl{https://proceedings.neurips.cc/paper/2020/file/1457c0d6bfcb4967418bfb8ac142f64a-Paper.pdf}.

\bibitem{rouge}
\selectlanguageifdefined{english}
\BibEmph{Lin~Chin-Yew}. {ROUGE}: A Package for Automatic Evaluation of
  Summaries~// Text Summarization Branches Out. \BibDash
\newblock Barcelona, Spain~: Association for Computational Linguistics, 2004.
  \BibDash Jul. \BibDash
\newblock P.~74--81. \BibDash
\newblock Access mode: \BibUrl{https://www.aclweb.org/anthology/W04-1013}.

\bibitem{linger2020batch}
\selectlanguageifdefined{english}
\BibEmph{Linger~Mathis, Hajaiej~Mhamed}. Batch Clustering for Multilingual News
  Streaming~// Proceedings of Text2Story - Third Workshop on Narrative
  Extraction From Texts co-located with 42nd European Conference on Information
  Retrieval, Text2Story@ECIR 2020, Lisbon, Portugal, April 14th, 2020 [online
  only]. \BibDash
\newblock Vol.~2593 of \BibEmph{{CEUR} Workshop Proceedings}. \BibDash
\newblock CEUR-WS.org, 2020. \BibDash
\newblock P.~55--61. \BibDash
\newblock Access mode: \BibUrl{http://ceur-ws.org/Vol-2593/paper7.pdf}.

\bibitem{hnsw}
\selectlanguageifdefined{english}
\BibEmph{Malkov~Yu~A., Yashunin~D.~A.} Efficient and Robust Approximate Nearest
  Neighbor Search Using Hierarchical Navigable Small World Graphs~//
  \href{http://dx.doi.org/10.1109/TPAMI.2018.2889473}{\BibEmph{IEEE
  Transactions on Pattern Analysis and Machine Intelligence}}. \BibDash
\newblock 2020. \BibDash
\newblock Vol.~42, no.~4. \BibDash
\newblock P.~824--836.

\bibitem{headline_gen}
\selectlanguageifdefined{english}
\BibEmph{Malykh~V.A.~Kalaidin~P.S.} Headline Generation Shared Task on
  Dialogue’2019~// Computational Linguistics and Intellectual Technologies,
  Papers from the Annual International Conference "Dialogue". \BibDash
\newblock 2019.

\bibitem{miranda-etal-2018-multilingual}
\selectlanguageifdefined{english}
\href{http://dx.doi.org/10.18653/v1/D18-1483}{Multilingual Clustering of
  Streaming News}~/ Sebasti{\~a}o~Miranda, Art{\=u}rs~Znoti{\c{n}}{\v{s}},
  Shay~B.~Cohen, Guntis~Barzdins~// Proceedings of the 2018 Conference on
  Empirical Methods in Natural Language Processing. \BibDash
\newblock Brussels, Belgium~: Association for Computational Linguistics, 2018.
  \BibDash Oct.-Nov. \BibDash
\newblock P.~4535--4544. \BibDash
\newblock Access mode: \BibUrl{https://www.aclweb.org/anthology/D18-1483}.

\bibitem{takase2016neural}
\selectlanguageifdefined{english}
Neural headline generation on abstract meaning representation~/ Sho~Takase,
  Jun~Suzuki, Naoaki~Okazaki et~al.~// Proceedings of the 2016 conference on
  empirical methods in natural language processing. \BibDash
\newblock 2016. \BibDash
\newblock P.~1054--1059.

\bibitem{rupnik}
\selectlanguageifdefined{english}
News across languages-cross-lingual document similarity and event tracking~/
  Jan~Rupnik, Andrej~Muhic, Gregor~Leban et~al.~// \BibEmph{Journal of
  Artificial Intelligence Research}. \BibDash
\newblock 2016. \BibDash
\newblock Vol.~55. \BibDash
\newblock P.~283--316.

\bibitem{rand}
\selectlanguageifdefined{english}
\BibEmph{Rand~William~M}. Objective criteria for the evaluation of clustering
  methods~// \BibEmph{Journal of the American Statistical association}.
  \BibDash
\newblock 1971. \BibDash
\newblock Vol.~66, no. 336. \BibDash
\newblock P.~846--850.

\bibitem{sentence_bert}
\selectlanguageifdefined{english}
\BibEmph{Reimers~Nils, Gurevych~Iryna}.
  \href{http://dx.doi.org/10.18653/v1/D19-1410}{Sentence-BERT: Sentence
  Embeddings using Siamese BERT-Networks}~// Proceedings of the 2019 Conference
  on Empirical Methods in Natural Language Processing and the 9th International
  Joint Conference on Natural Language Processing, {EMNLP-IJCNLP} 2019, Hong
  Kong, China, November 3-7, 2019. \BibDash
\newblock Association for Computational Linguistics, 2019. \BibDash
\newblock P.~3980--3990. \BibDash
\newblock Access mode: \BibUrl{https://doi.org/10.18653/v1/D19-1410}.

\bibitem{participants136}
\selectlanguageifdefined{english}
\BibEmph{Smirnova, Vatolin, Shkarin}. {Russian News Similarity Detection with
  SBERT: pre-training and fine-tuning}~// Computational Linguistics and
  Intellectual Technologies: Papers from the Annual Conference ``Dialogue''.
  \BibDash
\newblock Vol.~XX. \BibDash
\newblock 2021. \BibDash
\newblock P.~xx--xx.

\bibitem{stankevicius}
\selectlanguageifdefined{english}
\BibEmph{Stankevicius~Lukas, Lukosevicius~Mantas}. Testing Pre-trained
  Transformer Models for Lithuanian News Clustering~// Proceedings of the
  Information Society and University Studies 2020, Kaunas, Lithuania, April 23,
  2020. \BibDash
\newblock Vol.~2698 of \BibEmph{{CEUR} Workshop Proceedings}. \BibDash
\newblock CEUR-WS.org, 2020. \BibDash
\newblock P.~46--53. \BibDash
\newblock Access mode: \BibUrl{http://ceur-ws.org/Vol-2698/p08.pdf}.

\bibitem{chakraborty2016stop}
\selectlanguageifdefined{english}
Stop clickbait: Detecting and preventing clickbaits in online news media~/
  Abhijnan~Chakraborty, Bhargavi~Paranjape, Sourya~Kakarla, Niloy~Ganguly~//
  2016 IEEE/ACM international conference on advances in social networks
  analysis and mining (ASONAM)~/ IEEE. \BibDash
\newblock 2016. \BibDash
\newblock P.~9--16.

\bibitem{seq2seq}
\selectlanguageifdefined{english}
\BibEmph{Sutskever~Ilya, Vinyals~Oriol, Le~Quoc~V}. Sequence to Sequence
  Learning with Neural Networks~// Advances in Neural Information Processing
  Systems. \BibDash
\newblock Vol.~27. \BibDash
\newblock Curran Associates, Inc., 2014. \BibDash
\newblock Access mode:
  \BibUrl{https://proceedings.neurips.cc/paper/2014/file/a14ac55a4f27472c5d894ec1c3c743d2-Paper.pdf}.

\bibitem{allan1998topic}
\selectlanguageifdefined{english}
Topic Detection and Tracking Pilot Study: Final Report~/ J.~Allan,
  J.~Carbonell, G.~Doddington et~al.~// Proceedings of the DARPA Broadcast News
  Transcription and Understanding Workshop. \BibDash
\newblock Lansdowne, VA, USA, 1998. \BibDash Feb. \BibDash
\newblock P.~194--218. \BibDash
\newblock 007.

\bibitem{use}
\selectlanguageifdefined{english}
Universal sentence encoder~/ Daniel~Cer, Yinfei~Yang, Sheng-yi~Kong et~al.~//
  \BibEmph{arXiv preprint arXiv:1803.11175}. \BibDash
\newblock 2018.

\bibitem{xlm_r}
\selectlanguageifdefined{english}
Unsupervised Cross-lingual Representation Learning at Scale~/ Alexis~Conneau,
  Kartikay~Khandelwal, Naman~Goyal et~al.~// \BibEmph{CoRR}. \BibDash
\newblock 2019. \BibDash
\newblock Vol. abs/1911.02116. \BibDash
\newblock \href{http://arxiv.org/abs/1911.02116}{1911.02116}.

\bibitem{participants138}
\selectlanguageifdefined{english}
{Using Generative Pretrained Transformer-3 Models for Russian News Clustering
  and Title Generation tasks}~/ ~Tikhonova, ~Pisarevskaya, ~Shliazhko,
  ~Shavrina~// Computational Linguistics and Intellectual Technologies: Papers
  from the Annual Conference ``Dialogue''. \BibDash
\newblock Vol.~XX. \BibDash
\newblock 2021. \BibDash
\newblock P.~xx--xx.

\bibitem{voropaev}
\selectlanguageifdefined{english}
\BibEmph{Voropaev~Pavel, Sopilnyak~Olga}. {Comparison of news clustring
  methods}. \BibDash
\newblock 2020.

\bibitem{participants140}
\selectlanguageifdefined{english}
\BibEmph{Voropaev~Pavel, Sopilnyak~Olga}. {Transformer-based Embeddings for
  Russian News Clustering and Headline Selection}~// Computational Linguistics
  and Intellectual Technologies: Papers from the Annual Conference
  ``Dialogue''. \BibDash
\newblock Vol.~XX. \BibDash
\newblock 2021. \BibDash
\newblock P.~xx--xx.

\bibitem{zhang1996birch}
\selectlanguageifdefined{english}
\BibEmph{Zhang~Tian, Ramakrishnan~Raghu, Livny~Miron}. BIRCH: an efficient data
  clustering method for very large databases~// \BibEmph{ACM sigmod record}.
  \BibDash
\newblock 1996. \BibDash
\newblock Vol.~25, no.~2. \BibDash
\newblock P.~103--114.

\bibitem{mt5}
\selectlanguageifdefined{english}
mT5: {A} massively multilingual pre-trained text-to-text transformer~/
  Linting~Xue, Noah~Constant, Adam~Roberts et~al.~// \BibEmph{CoRR}. \BibDash
\newblock 2020. \BibDash
\newblock Vol. abs/2010.11934. \BibDash
\newblock \href{https://arxiv.org/abs/2010.11934}{2010.11934}.

\end{thebibliography}
\bibliographystyle{ugost2008ls}

\end{document}